\documentclass{article}
\usepackage{moreverb}

\usepackage{subfigure}
\usepackage{url}
\usepackage{amsfonts}
\usepackage{amssymb}
\usepackage{pseudocode}
\usepackage{fancybox}
\usepackage{amsthm}
\usepackage{latexsym}
\usepackage{xspace}
\usepackage{graphicx}
\usepackage{amsmath}
\usepackage[latin1]{inputenc}
\usepackage{epsfig}
\usepackage{mathrsfs}
\usepackage{xcolor}
\usepackage{listings}

\newtheorem{ex}{Example}[section]

\newcommand{\bone}{\boldsymbol{1}}

\newcommand{\bx}{\boldsymbol{x}}
\newcommand{\bw}{\boldsymbol{w}}
\newcommand{\by}{\boldsymbol{y}}

\usepackage{authblk}

\title{Soft Rule Ensembles for Supervised Learning}
\author[1]{Deniz Akdemir\thanks{da346@cornell.edu}}
\author[1,2]{Nicolas Heslot}
\author[1,3]{Jean-Luc Jannink}
\affil[1]{Bradfield Hall, Department of Plant Breeding \& Genetics 
 Cornell University, Ithaca, NY 14853, USA}
\affil[2]{Limagrain Europe, Chappes, France}
\affil[3]{USDA-ARS, Cornell University Ithaca, NY, USA}

\date{\today}
\begin{document}

\maketitle

\begin{abstract}
In this article supervised learning problems are solved using soft rule ensembles.  We first review the importance sampling learning ensembles (ISLE) approach that is useful for generating hard rules. The soft rules are obtained with logistic regression using the corresponding hard rules. Soft rules are useful when both the response and the input variables are continuous because the soft rules provide smooth transitions around the boundaries of a hard rule. Various examples and simulation results show that soft rule ensembles can improve predictive performance over hard rule ensembles.
\end{abstract}


\section{Introduction}

A relatively new approach to modeling data,  namely the ensemble learning (\cite{ho1990combination},  \cite{hansen1990neural},  \cite{kleinberg1990stochastic}) challenges the monist views by providing solutions to complex problems simultaneously from a number of models. By focusing on regularities and stable common behavior,  ensemble modeling approaches provide solutions that as a whole outperform the single models. Some influential early works in ensemble learning were by Breiman  with Bagging (bootstrap aggregating) (\cite{breiman1996bagging}), and Freund and Shapire  with AdaBoost (\cite{freund1996experiments}). All of these methods involve random sampling the ''space of models'' to produce an ensemble of models. 

Although not necessary, in practice ensemble models are usually used with regression / classification trees or binary rules extracted from them which are discontinuous and piecewise constant. In order to approximate a smooth response variable, a large number of trees or rules with many splits are needed.  This causes data fragmentation for high dimensional problems where the data is already sparse. The soft rule ensembles that are proposed in this paper attacks this problem by replacing the hard rules with smooth functions.    

In the rest of this section, we review the ensemble model generation and post-processing approach due to Popescu \& Friedman (\cite{friedman2003importance}). Their approach attempts to unify ensemble learning methods. In the next section, we explain how to convert hard rules to soft rules using bias corrected logistic regression. In section 3, we compare the soft and hard rule examples. The paper concludes with our comments and discussions. 

Suppose we are asked to predict the continuous outcome variable $y$ from $p$ vector of input variables $\bx.$ We restrict the prediction models to the model family $\mathscr{F}=\{f(\bx; \theta): \theta \in \Theta\}.$ The models considered by  the ISLE framework  have an additive expansion of the form: 
\begin{equation}F(\bx)=w_0+\sum_{j=1}^{M}w_{j} f(\bx, \theta_j)\label{eq:additivemodel}\end{equation}
where $\{f(\bx, \theta_j)\}_{j=1}^{M}$ are base learners selected from $\mathscr{F}.$ Popescu \& Friedman's ISLE approach (\cite{friedman2003importance}) uses a heuristic two-step approach to arrive at $F(\bx)$. The first step involves sampling the space of possible models to obtain $\{\widehat{\theta}_j\}_{j=1}^{M}$. The models in the model family $\mathscr{F}$ are sampled using perturbation sampling; by varying case weights, data values,  variable subsets, or partitions of the input space (\cite{seni2010ensemble}).  The second step combines the predictions from these models by choosing weights $\{w_j\}_{j=0}^{M}$ in (\ref{eq:additivemodel}).

The pseudo code to produce $M$ models $\{f(\bx, \widehat{\theta}_j)\}_{j=1}^{M}$ under ISLE framework is given below:
\begin{pseudocode}{ISLE}{M, v, \eta}
\label{ISLE}
$$F_0(\bx)=0.$$ \\
\FOR $j=1$ \TO $M$ \DO
\BEGIN
$$(\widehat{c}_j, \widehat{\theta}_j)= \underset{(c,\theta)}{\operatorname{argmin}}\sum_{i \in S_j(\eta)} L(y_i, F_{j-1}(\bx_i)+cf(\bx_i, \theta))$$ \\
$$T_j(\bx)=f(\bx, \widehat{\theta}_j)$$ \\
$$F_j(\bx)=F_{j-1}(\bx)+\nu\widehat{c}_j T_j(\bx)$$ \\
\END \\
\RETURN{$$\{T_j(\bx)\}_{j=1}^M$$ and $$F_M(\bx).$$}
\end{pseudocode}

Here, $L(.,.)$ is a loss function, $S_j(\eta)$ is a subset of the indices $\{1,2,\ldots, n\}$ chosen by a sampling scheme $\eta,$ $0\leq \nu \leq 1$ is a memory parameter. 

The classic ensemble methods of Bagging, Random Forest, AdaBoost, and Gradient Boosting are special cases of the generic ensemble generation procedure (\cite{seni2010ensemble}). The weights $\{w_j\}_{j=0}^{M}$ can be selected in a number of ways, for Bagging and Random Forests these weights are set to predetermined values, i.e. $w_0=0$ and $w_j=\frac{1}{M}$ for $j=1,2,\ldots,M.$ Boosting calculates these weights in stage wise fashion at each step by having positive memory $\mu,$ estimating $c_j$ and takes $F_M(\bx)$ as the final prediction model.  

Friedman \& Popescu (\cite{friedman2003importance}) recommend learning the weights $\{w_j\}_{j=0}^{M}$ using LASSO (\cite{tibshirani1996regression}). Let $T={\left(T_j(\bx_i) \right)_{i=1}^n}_{m=1}^M$ be the $n\times M$ matrix of predictions for the $n$ observations by the $M$ models in an ensemble. The weights $(w_0,\bw=\{w_m\}_{m=1}^{M})$ are obtained from \begin{equation}\hat{\bw}=\underset{\bw}{\operatorname{argmin}} (\by-w_0\bone_n-T\bw)'(\by-w_0\bone_n-T\bw)+\lambda \sum_{j=1}^M|w_m|.\end{equation}
$\lambda>0$ is the shrinkage operator, larger values of $\lambda$ decreases the number of models included in the final prediction model.  The final ensemble model is given by $F(\bx)=w_0+\sum_{m=1}^{M}w_{m} T_m(\bx).$

The base learners in the preceding sections of this article can be of any kind, however usually they are regression or decision trees. Each decision tree in the ensemble partitions the input space using the product of indicator functions of ''simple'' regions based on several input variables. A tree with $K$ terminal nodes define a $K$ partition of the input space where the membership to a specific node, say node $k,$ can be accomplished by applying the conjunctive rule $r_k(\bx)=\prod_{l=1}^{p}I(x_l\in s_{lk}),$ where $I(.)$ is the indicator function. The regions $s_{lk}$ are intervals for  continuous variables and  subsets of the levels for  categorical variables. Therefore, a rule corresponds to a region that is the intersection of half spaces defined by hyperplanes that are orthogonal to the axis of the predictor variables.

A regression tree with $K$ terminal nodes can be written as \begin{equation}T(\bx)=\sum_{k=1}^{K}\beta_k r_k(\bx).\end{equation} Trees with many terminal nodes usually produce more complex rules and tree size is an important meta-parameter which we can control by maximum tree depth and cost pruning. 

\begin{figure}[htbp]
	\centering
		\includegraphics[width=.7\textwidth]{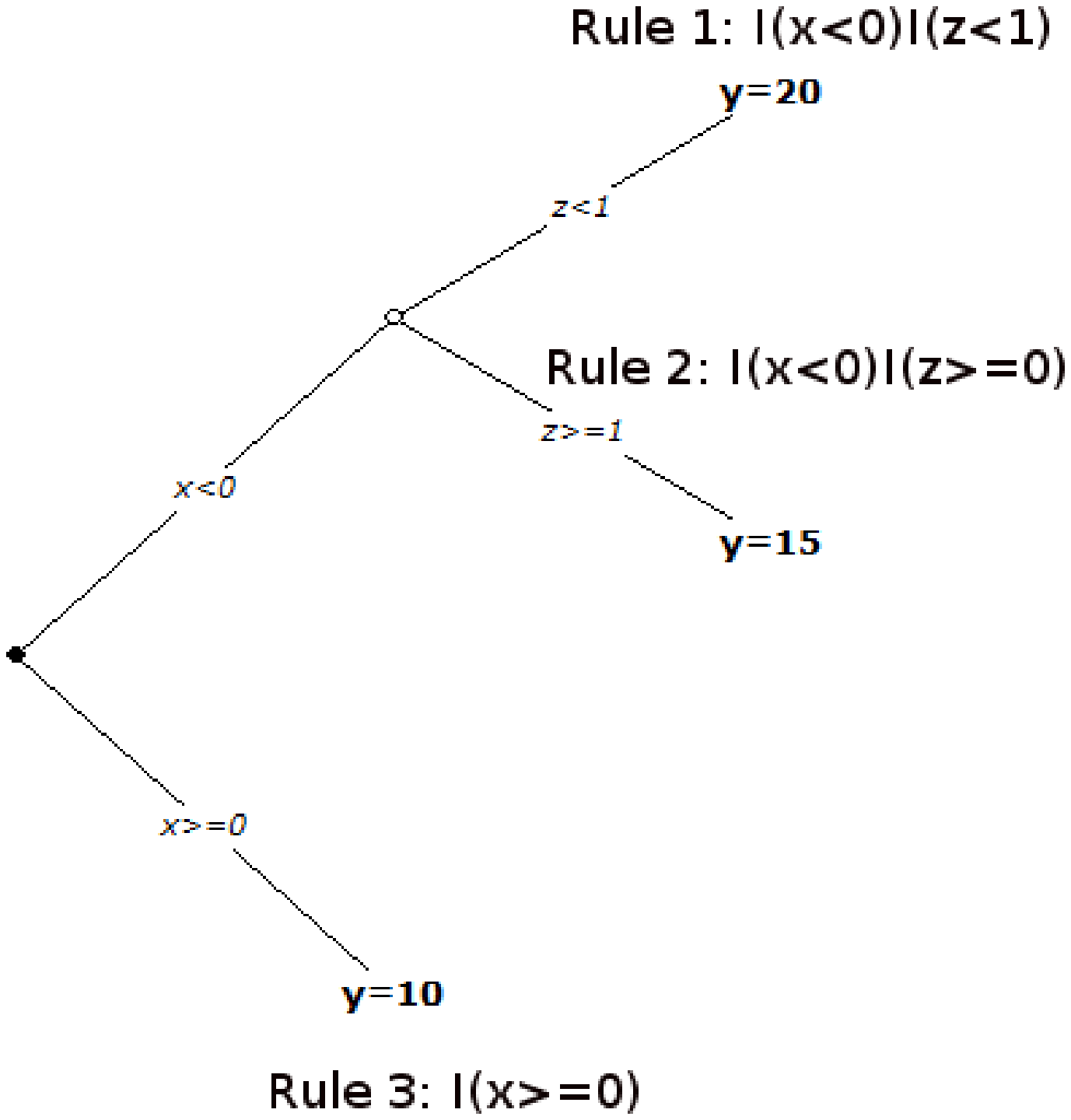}
	\label{fig:treetorules}
\end{figure}

Given a set of decision trees, rules can be extracted from each of these trees to define a collection of conjunctive rules (Figure \ref{fig:treetorules}). A conjunctive rule $r(\bx)=\prod_{l=1}^{p}I(x_l\in s_{l})$ can also be expressed as a logic rule (also called Boolean expressions and logic statement) involving  only the $\wedge$ (''and'') operator. In general, a logic statement is constructed using the operators $\wedge$ (''and''), $\vee$ (''or'') and $ ^c$ (''not'') and brackets. An example simple logic rule is $l(\bx)=[I(x_1\in s_{1})\vee I^c(x_2\in s_{2})]\wedge I(x_3\in s_{3}).$ Logic Regression (\cite{kooperberg2005identifying}) is an adaptive regression methodology that constructs logic rules from binary input variables. Simple conjunctive rules that are learned by the the ISLE approach can be used as input variables to logic regression to combine these rules. However, the representation of a logic rule in general is not unique and it can be shown that all logic rules can be expressed in disjunctive normal form where we only use $\vee$ combinations of $\wedge$ (not necessarily simple) terms.  

Let  $R={\left(r_k(\bx_i) \right)_{i=1}^n}_{l=1}^L$ be the $n\times L$ matrix of rules for the $n$ observations by the $L$ rules in the ensemble. The \bfseries rulefit \normalfont algorithm of Friedman \& Popescu \cite{friedman2008predictive} uses the weights $(w_0,\bw=\{ {w_l}\}_{l=1}^{L})$ that are estimated from \begin{equation}\hat{\bw}=\underset{\bw}{\operatorname{argmin}} (\by-w_0\bone_n-R\bw)'(\by-w_0\bone_n-R\bw)+\lambda \sum_{l=1}^L|w_l|\end{equation} in the final prediction model $F(\bx)=w_0+\sum_{l=1}^{L}w_{l} r_l(\bx).$

Rule ensembles are shown to produce improved accuracy over traditional ensemble methods like random forests, bagging, boosting and ISLE (\cite{friedman2003importance}, \cite{friedman2008predictive} and \cite{seni2010ensemble}).
  
\section{Soft Rules from Hard Rules}

Soft rules which take values in $[0,1]$ are obtained by replacing each hard rule $r(\bx)$ with a logistic function of the form \[s(\bx)=\frac{1}{1+exp(-g(\bx;\theta))}.\] The value of a soft rule $s(\bx)$ can be viewed as the probability that that rule is fired for $\bx.$  

In this paper, $g(\bx;\theta)$ includes additive terms of order two without any interaction terms in the variables which were used explicitly in the construction of the rule $r(\bx).$ The model is built using best subsets regression, selecting the best subset of terms for which the AIC is minimized. The coefficients $\theta$ of the function $g(\bx;\theta)$ are to be estimated from the examples of $\bx$ and $r(\bx)$ in the training data.

A common problem with logistic regression is the problem of (perfect) separation (\cite{heinze2002solution}) which occurs when the response variable can be perfectly separated by one or a linear combination of a few explanatory variables. When this is the case, the likelihood function becomes monotone and non finite estimates of coefficients are produced. In order to deal with the problem of separation, Firth's bias corrected likelihood approach (\cite{firth1993bias}) has been recommended (\cite{heinze2002solution}). The bias corrected likelihood approach to logistic regression are guaranteed to produce finite  estimates and  standard errors.

Maximum likelihood estimators of the coefficients $\theta$ are obtained as the solution to the score equation $$d log L(\theta)/ d \theta =U(\theta)=0$$ where $L(\theta)$ is the likelihood function. Firth's bias corrected likelihood uses a modified likelihood function $$L^*(\theta)=L(\theta)|i(\theta)|^{1/2}$$ where $i(\theta)$ is the Jeffreys (\cite{jeffreys1946invariant}) invariant prior, the Fisher information matrix.


Using the modified likelihood function the score function for the logistic model is given by $U^*(\theta)=(U^*(\theta_1),U^*(\theta_2),\ldots,U^*(\theta_k))'$ where $$U^*(\theta_j)=\sum_{i=1}^{n}\{r(\bx_i)-g(\bx_i;\theta)+h_i(\frac{1}{2}-g(\bx_i;\theta))\}\frac{\partial g(\bx_i;\theta)}{\partial \theta_j}$$ for $j=1,2,\ldots,k$ and $k$ is the number of coefficients in $g(\bx;\theta).$ Here, $h_i$ for $i=1, 2,\ldots, n$ are the $i$th diagonal elements of the hat matrix $$H=W^{1/2}X(X'WX)^{-1}X'W^{1/2}$$ and $W=diag\{g(\bx_i;\theta)(1-g(\bx_i;\theta)).\}$  Bias corrected estimates can be obtained in an iterative fashion using $$\theta^{t+1}=\theta^{t}+I^{-1}(\theta^{t})U^*(\theta^{t}).$$ Our programs utilize the  ''brglm'' package in R (\cite{kosmidis2008brglm}) that fits binomial-response generalized linear models using the bias-reduction.

In \cite{da2008tree} hard rules are replaced with products of univariate logistic functions to build the models called tree-structured smooth transition regression models that generalize the regression tree models. A similar model called soft decision trees are introduced in \cite{irsoy2012soft}. These authors use logistic functions to calculate gating probabilities at each node in an hierarchical structure where the children of each node are selected with a certain probability, the terminal nodes  of the incrementally learned trees are represented as a product of logistic functions. In \cite{dvovrak2007softening} tree splits are softened using simulated annealing. Fuzzy decision trees were presented by \cite{olaru2003complete}. Perhaps, these models can be used to generate soft rules. However, in this paper, we use a simpler approach where we utilize logistic functions only at the terminal nodes of the trees built by the CART algorithm.

\begin{figure}[htbp]
	\centering
		\includegraphics[width=1.00\textwidth]{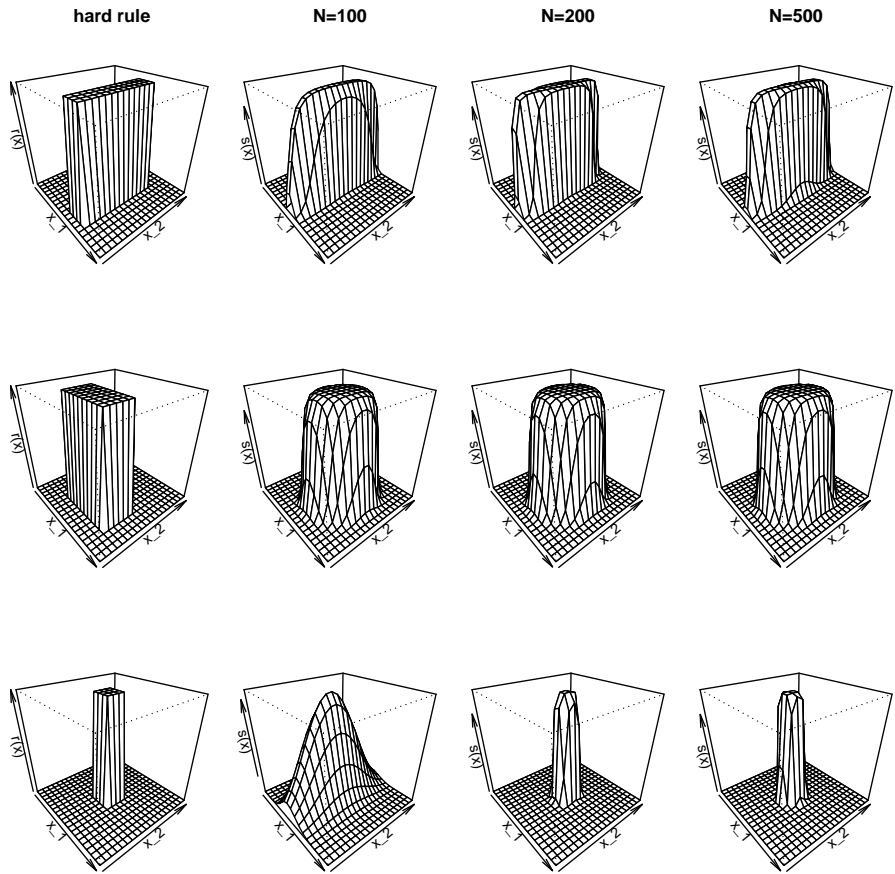}
	\caption{Hard rules and the corresponding soft rules estimated from the training data. It is clear that the soft rules provide a smooth approximation to the hard rules. 
}
	\label{contourplot}
\end{figure}

In Figure \ref{contourplot}, we present simple hard rules and the corresponding soft rules estimated from the training data. It is clear that the soft rules provide a smooth approximation to the hard rules. 

Let $R={\left(r_l(\bx_i) \right)_{i=1}^n}_{l=1}^L$ be the $n\times L$ matrix of  $L$ rules for the $n$ observations in the training sample. Letting $s_l(\bx;\hat{\theta}_l)$ be the soft rule corresponding to the $l$th hard rule, define $S={\left(s_l(\bx_i) \right)_{i=1}^n}_{l=1}^L$ as the $n\times L$ matrix of $L$ soft rules for the $n$ observations. 

The weights for the soft rules can be estimated from the LASSO: \begin{equation}\label{rulefitlasso}\hat{\bw}=\underset{\bw}{\operatorname{argmin}} (\by-w_0\bone_n-S\bw)'(\by-w_0\bone_n-S\bw)+\lambda \sum_{l=1}^L|w_l|.\end{equation} 

This leads to the final soft rule ensemble prediction model $F(\bx)=\hat{w}_0+\sum_{l=1}^{L}\hat{w}_{l} s_l(\bx).$
  
The only additional step for building our soft rules ensembles is the hard to soft rule conversion step. This  makes our algorithm slower than the the rulefit algorithm (\cite{friedman2003importance}), at times, $10$ times slower.  However, we have implemented our soft rules algorithm in the R language and successfully applied it to several high dimensional problems. We expect that a faster implementation is possible if the code is migrated to a faster programming language. In addition parts or the whole of the soft rule generation process can be accomplished by parallel processing. 

For completeness and easy reference, we summarize the steps for soft rule ensemble generation and fitting:
\begin{enumerate}
\item Use ISLE algorithm to generate M trees: T(X).
\item Extract hard rules from T(X): R(X).
\item Convert hard rules to soft rules: S(X).
\item Obtain soft rule weights by LASSO. 
\end{enumerate}

We should note that no additional meta-parameters are introduced to produce soft rules from hard rules. We set these meta-parameters along the lines  of the recommendations in previous work (\cite{friedman2003importance}, \cite{friedman2008predictive} and \cite{seni2010ensemble}).

There are several fast algorithms that can accomplish the LASSO post processing for large datasets (large n or p): Recent pathwise coordinate descent (\cite{friedman2007pathwise}) (implemented in ''glmnet'' with R (\cite{friedman2009glmnet})) algorithm provide the entire path of solutions. We have used the ''glmnet'' in our illustrations in the next section, the value of the sparsity parameter was set by minimizing the mean cross-validated error. Due to the well known selection property of the lasso penalty, only $10\%$-$20\%$ of the rules are retained in the final model after the post-processing step.

\section{Illustrations}

In this section we are going to compare the soft rule and hard rule ensembles. The prediction accuracy is taken as the cross validated correlation or the mean square error (MSE) calculated for the predicted and true target variable values for regression examples and cross validated area under the ROC curve for classification examples.

The number of trees to be generated by the ISLE algorithm was set to $M=400,$ each tree used $30\%$ of of the individuals and $10\%$ of the input variables randomly selected from the training set. Larger trees allow more complex rules to be produced and therefore controlling tree size controls the maximum rule complexity. A choice of this parameter can be based on prior knowledge or based on experiments with different values. We have tried differing three depths for obtaining models with varying complexity. For most of the examples accuracies were reported for each tree depth. In example \ref{explantbreeding}, models with differing rule depths were trained and only the models with best cross validated performances were reported. In addition, the memory parameter $\nu$ of the ISLE ensemble generation algorithm is set to zero in all the following examples. 

\begin{ex}(Boston Housing Data, Regression) In order to compare the performance of prediction models based on hard and soft rules we use the famous benchmark ''Boston Housing''  data set (\cite{harrison1978hedonic}). This data set includes n=506 observations and p=14 variables. The response variable is the median house value from the rest of the 13 variables in the data set. 10 fold cross validated accuracies are displayed in Table \ref{tab:boston}. Using soft rules we gain couple points improvement on the accuracies. 

\begin{table}[htbp]
\begin{center}
\begin{tabular}{|r|r|r|r|r|}
\hline
\multicolumn{1}{|l|}{} & \multicolumn{1}{l|}{accuracy} & \multicolumn{1}{l|}{} & \multicolumn{1}{l|}{rmse} & \multicolumn{1}{l|}{} \\ \hline
\multicolumn{1}{|l|}{tree size} & \multicolumn{1}{l|}{hard} & \multicolumn{1}{l|}{soft} & \multicolumn{1}{l|}{hard} & \multicolumn{1}{l|}{soft} \\ \hline
2 & 0.91 & 0.92 & 3.78 & 3.60 \\ \hline
3 & 0.93 & 0.93 & 3.40 & 3.42 \\ \hline
4 & 0.94 & 0.93 & 3.18 & 3.29 \\ \hline
5 & 0.93 & 0.94 & 3.33 & 3.18 \\ \hline
\end{tabular}
\end{center}
\caption{The 10-fold cross validated prediction accuracies as measured by the correlation of the true and predicted values are given for the ''Boston housing data''.}
\label{tab:boston}
\end{table}
	
\end{ex}

As observed from the previous example, for continuous input variables soft rule ensembles might have better prediction accuracy then its hard rule ensembles counterpart. However, for problems with only categorical or discrete input variables, we do not expect to see the same improvements. The following example only uses discrete SNP markers (biallelic markers values coded as -1,0 and 1) as input variables and hard rules and soft rules give approximately the same accuracies. 
 
\begin{ex}\label{explantbreeding}(Plant Breeding Data, Regression) In our second example we analyze plant breeding data and compare the predictive performance of hard rules with soft rules. In both cases, the objective is to predict a quantitative trait (observed performance) using molecular markers data providing information about the genotypes of the plants. Predictions of phenotypes using numerous molecular markers at the same time is called genomic selection and has received lately a lot of attention in the plant and animal breeding communities. Rule ensembles used with genetic marker data implicitly captures epistasis (interaction between markers) in a highly dimensional context while retaining interpretability of the model. 

The first data set (Bay x Sha) contains measurements on flowering time under short day length, dry matter under non limiting or limiting conditions from 422 recombinant inbred lines from a biparental population of Arabidopsis thaliana plants from 2 ecotypes, Bay-0 and Shadara genotyped with 69 SSRs. Data available from the Study of the Natural Variation of A. thaliana website (\cite{heslot2012genomic}, \cite{lorenzana2009accuracy}). 

The second data set (Wheat CIMMYT) is composed of 599 spring wheat inbred lines evaluated for yield in 4 different target environments (YLD1-YLD4). 1279 DArT markers were available for the 599 lines in the study (\cite{crossa2010prediction}).  The results are displayed in Table \ref{table:plant}.
\begin{table}[htbp]
\begin{center}
\begin{tabular}{|l|r|r|r|r|l|r|r|r|r|}
\hline
cimmyt & \multicolumn{1}{l|}{} & \multicolumn{1}{l|}{} & \multicolumn{1}{l|}{} & \multicolumn{1}{l|}{} & BaySha & \multicolumn{1}{l|}{} & \multicolumn{1}{l|}{} & \multicolumn{1}{l|}{} & \multicolumn{1}{l|}{} \\ \hline
 & \multicolumn{1}{l|}{accuracy} & \multicolumn{1}{l|}{} & \multicolumn{1}{l|}{rmse} & \multicolumn{1}{l|}{} &  & \multicolumn{1}{l|}{accuracy} & \multicolumn{1}{l|}{} & \multicolumn{1}{l|}{rmse} & \multicolumn{1}{l|}{} \\ \hline
tree size & \multicolumn{1}{l|}{hard} & \multicolumn{1}{l|}{soft} & \multicolumn{1}{l|}{hard} & \multicolumn{1}{l|}{soft} & tree size & \multicolumn{1}{l|}{hard} & \multicolumn{1}{l|}{soft} & \multicolumn{1}{l|}{hard} & \multicolumn{1}{l|}{soft} \\ \hline
\multicolumn{1}{|r|}{2} & 0.50 & 0.51 & 0.87 & 0.86 & \multicolumn{1}{r|}{2} & 0.86 & 0.86 & 4.66 & 4.64 \\ \hline
 & 0.41 & 0.41 & 0.92 & 0.92 &  & 0.67 & 0.67 & 2.17 & 2.18 \\ \hline
 & 0.36 & 0.36 & 0.96 & 0.96 &  & 0.37 & 0.37 & 1.18 & 1.18 \\ \hline
 & 0.42 & 0.42 & 0.92 & 0.92 & \multicolumn{1}{r|}{3} & 0.86 & 0.85 & 4.75 & 4.77 \\ \hline
\multicolumn{1}{|r|}{3} & 0.52 & 0.52 & 0.86 & 0.86 &  & 0.62 & 0.62 & 2.30 & 2.30 \\ \hline
 & 0.42 & 0.43 & 0.92 & 0.91 &  & 0.33 & 0.33 & 1.22 & 1.22 \\ \hline
 & 0.40 & 0.40 & 0.93 & 0.93 &  & \multicolumn{1}{l|}{} & \multicolumn{1}{l|}{} & \multicolumn{1}{l|}{} & \multicolumn{1}{l|}{} \\ \hline
 & 0.40 & 0.41 & 0.94 & 0.93 &  & \multicolumn{1}{l|}{} & \multicolumn{1}{l|}{} & \multicolumn{1}{l|}{} & \multicolumn{1}{l|}{} \\ \hline
\end{tabular}
\end{center}
\caption{Accuracies of hard and soft rule ensembles are compared by the cross validated Pearson's correlation coefficients between the estimated and true values. For each data set we have trained models based on rules of depth 1 to 6,  the results from the model with best cross validated accuracies is reported. The results show no significant difference between hard or soft rules.}
\label{table:plant}
\end{table}

\end{ex}

When the task is classification, no significant improvement is achieved by preferring soft rules over hard rules. To compare soft and hard rule ensembles in the context of classification, we use the Arcene and Madelon datasets downloaded from UCI Machine Learning Repository.
 
\begin{ex}(Arcene data, Classification) The task in arcene data is to classify patterns as normal or cancer based on mass-spectrometric data. There were 7000 initial input variables, 3000 probe input variables were added to increase the difficulty of the problem. There were 200 individuals in the data set. Areas under the ROC curves for soft and hard rule ensemble models based on rules with depths 2 to 8 are compared in Table \ref{roc} (left).
\end{ex}

\begin{ex}(Madelon data, Classification) Madelon is contains data points grouped in 32 clusters placed on the vertices of a five dimensional hypercube and randomly labeled +1 or -1. There were 500 continuous input variables, 480 of these were probes. There were 2600 labeled examples. Areas under the ROC curves for rules with depths 2 to 7 are compared in Table \ref{roc} (right).

\end{ex}\begin{table}[htbp]
\begin{center}
\begin{tabular}{|r|r|r|r|r|r|}
\hline
\multicolumn{1}{|l|}{Arcene} & \multicolumn{1}{l|}{} & \multicolumn{1}{l|}{} & \multicolumn{1}{l|}{Madelon} & \multicolumn{1}{l|}{} & \multicolumn{1}{l|}{} \\ \hline
\multicolumn{1}{|l|}{tree size} & \multicolumn{1}{l|}{hard} & \multicolumn{1}{l|}{soft} & \multicolumn{1}{l|}{tree size} & \multicolumn{1}{l|}{hard} & \multicolumn{1}{l|}{soft} \\ \hline
2 & 0.82 & 0.82 & 2 & 0.83 & 0.87 \\ \hline
3 & 0.87 & 0.82 & 3 & 0.86 & 0.88 \\ \hline
4 & 0.79 & 0.80 & 4 & 0.90 & 0.92 \\ \hline
\end{tabular}
\end{center}
\caption{10- fold cross validated areas under the ROC curves for soft and hard rule ensemble models based on rules with depths 2 to 7 for the Arsene (left) and Madelon (right) data sets. We do not observe any significant difference between soft and hard rules.}
\label{roc}
\end{table}

When both response and input variables are continuous, soft rules perform better than hard rules. In the last two examples, we compare our models via mean squared errors.  

\begin{figure}
\centering
\mbox{\subfigure{\includegraphics[width=2.5in]{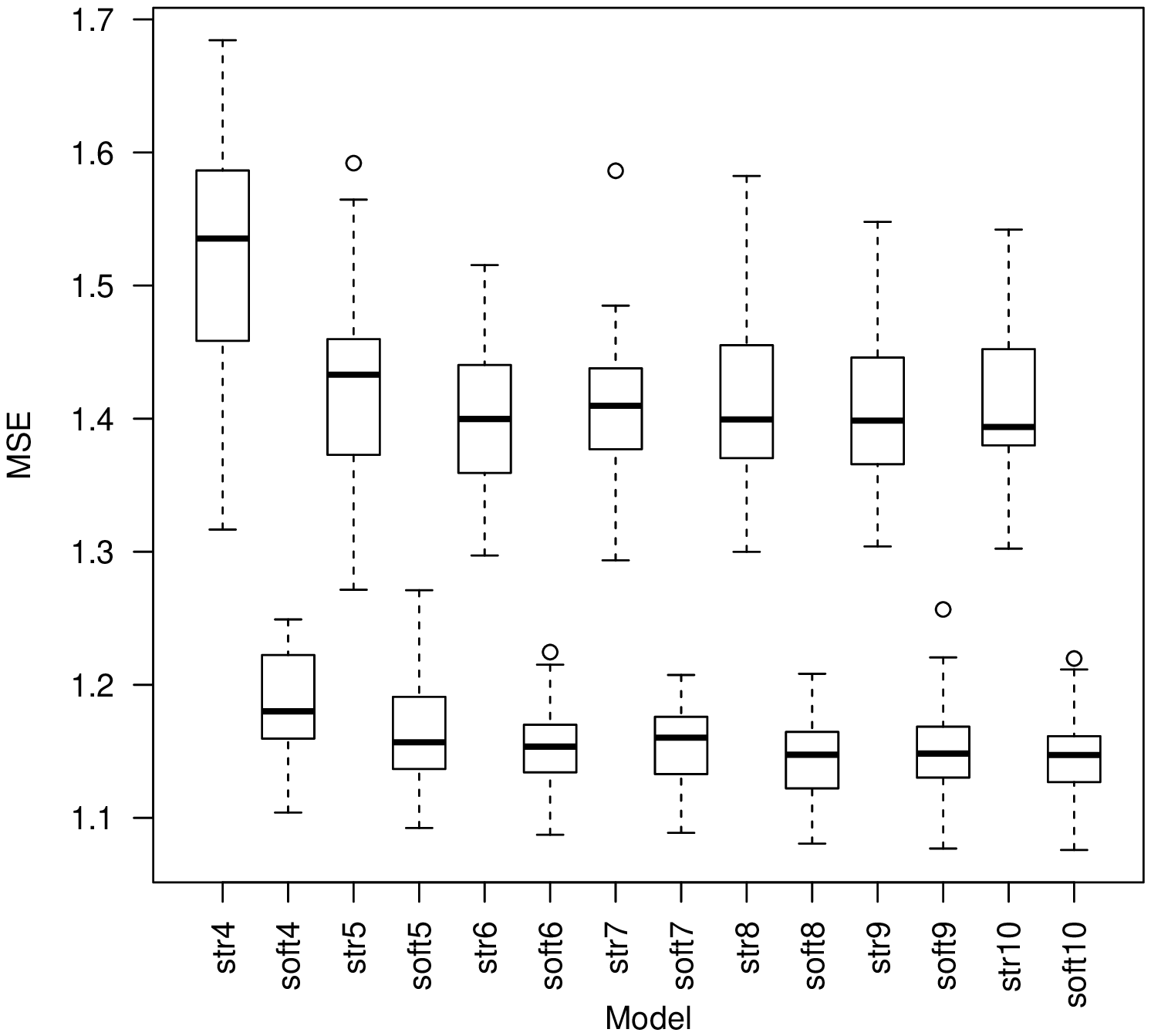}}
\quad
\subfigure{\includegraphics[width=2.5in]{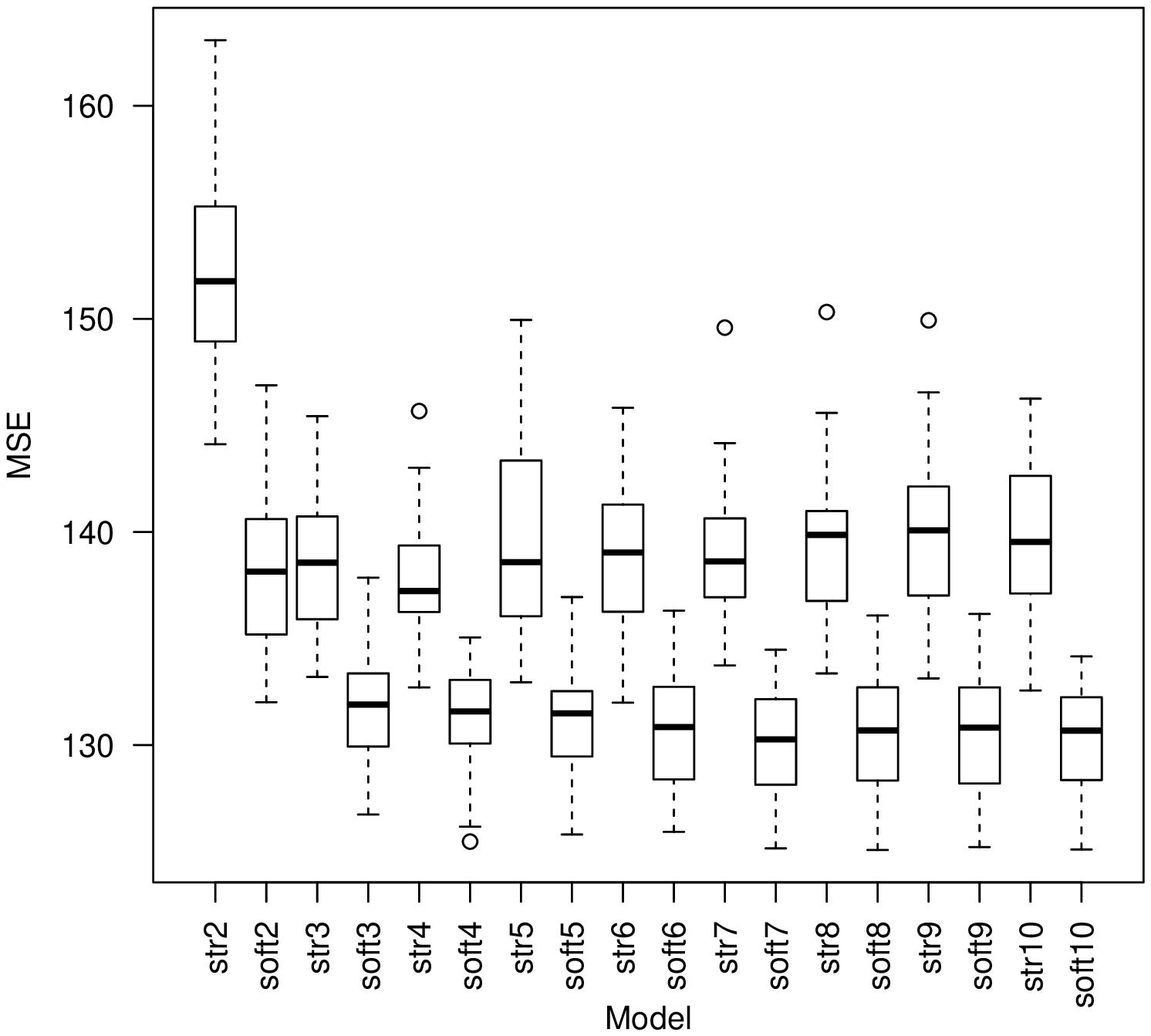}}}
\caption{The boxplots  summarize the prediction accuracies for soft and hard rules over 30 replications of the experiments described in Examples 3.5 and 3.6. In terms of mean squared errors the soft rule ensembles perform better than the corresponding hard rule ensembles for all values of tree depth. The hard rule ensemble models are denoted by ''str'' and soft rule ensemble models are denoted by ''soft''. The numbers next to these acronyms is the depth of the corresponding hard rules.}
\label{figuresim}
\end{figure}

\begin{ex} (Simulated Data, Regression) This regression problem is described in Friedman (\cite{friedman1991multivariate}) and Breiman (\cite{breiman1996bagging}). Elements of the input vector $\bx=(x_1,x_2,\ldots,x_{10})$ are generated from $uniform(0,1)$ distribution independently, only 5 out of these 10 are actually used to calculate the target variable $y$ as $$y = 10 \sin(\pi x_1x_2) + 20(x_3-0.5)^2+ 10x_4 + 5x_5 + e$$ where $e\sim N(0,1).$ 1000 independent realizations of $(\bx,y)$ constitute the training data. Mean squared errors for models are calculated on a test sample of the same size. The boxplots in left Figure \ref{figuresim} summarize the prediction accuracies for soft and hard rules over 30 replications of the experiment.  

\end{ex}

\begin{ex} (Simulated Data, Regression)
Another problem described in Friedman (\cite{friedman1991multivariate}) and Breiman (\cite{breiman1996bagging}). Inputs
are 4 independent variables uniformly distributed over the ranges $0\leq x_1 \leq 100,$
$40\pi\leq x_2 \leq 560\pi,$ $0 \leq x_3 \leq 1,$ $1\leq x_4\leq 11.$ The outputs are created according to the formula $y = (x_1^2+ (x_2x_3-(1/(x_2x_4))^2)^{0.5}+e$ where $e$ is $N(0,sd=125)$. 1000 independent realizations of $(\bx,y)$ constitute the training data. Mean squared errors for models are calculated on a test sample of the same size. The boxplots in right Figure \ref{figuresim} summarize the prediction accuracies for soft and hard rules.  
\end{ex}

\section{Discussions}

As our examples in the previous section suggest, the best case for soft rules is when both input and output variables are continuous. For data sets with mixed input variables it might be better to use two sets of rules (hard rules for categorical variables and soft rules for  numerical variables) and combine in a supervised learning model with group lasso model (\cite{grouplasso}) or perhaps these rules can be combined in a multiple kernel model (\cite{bach1}, \cite{gonen2011multiple}).

The hard rules or the soft rules can be used as input variables in any supervised  or unsupervised learning problem. In \cite{akdemir2011ensemble}, several promising hard rule ensemble methods were proposed for supervised, semi-supervised and unsupervised learning. For instance, the model weights can be obtained using partial least squares regression.  A similarity matrix obtained from hard rules can be used as a learned kernel matrix in Gaussian process regression. It is straightforward to use these and similar methods with soft rules.   

Several model interpretation tools have been developed to use with rule ensembles and the ISLE models. These include local and global rule, variable and interaction importance measures and partial dependence functions (\cite{friedman2008predictive}). We can use the same tools to interpret the soft rule ensemble model. For example, the absolute values of the standardized coefficients can be used to evaluate the importance of rules. A measure of importance for each input variable can be obtained as the sum the importances of rules that involve that variable. The interaction importance measures and the partial dependence functions  described in \cite{friedman2008predictive} are general measures and they also naturally apply to our soft rule ensembles.

The ensemble approaches are also a remedy for memory problems faced in analyzing big datasets. By adjusting the sampling scheme in the ISLE algorithm we were able to analyze large data sets which have thousands of variables and tens of thousands of individuals by only loading fractions of the data into the memory at a time. 

\section*{Acknowledgements}
This research was supported by the USDA-NIFA-AFRI Triticeae Coordinated Agricultural Project, award number 2011-68002-30029.

\bibliographystyle{plain}
\bibliography{ensemblebib}
\end{document}